\documentclass[sn-mathphys]{sn-jnl}% Math and Physical Sciences Reference Style
\jyear{2021}%

\theoremstyle{thmstyleone}%
\theoremstyle{thmstyletwo}%

\theoremstyle{thmstylethree}%

\usepackage[capitalize]{cleveref}

\begin{document}

\title[Fully $1\times1$ Convolutional Network for Lightweight Image Super-Resolution]{Fully $1\times1$ Convolutional Network for Lightweight Image Super-Resolution}

%%=============================================================%%
%% Prefix   -> \pfx{Dr}
%% GivenName    -> \fnm{Joergen W.}
%% Particle -> \spfx{van der} -> surname prefix
%% FamilyName   -> \sur{Ploeg}
%% Suffix   -> \sfx{IV}
%% NatureName   -> \tanm{Poet Laureate} -> Title after name
%% Degrees  -> \dgr{MSc, PhD}

%%%%%% Affiliations %%%%%%
\author[1]{\fnm{Gang} \sur{Wu}}
\author*[1]{\fnm{Junjun} \sur{Jiang}}
\email{jiangjunjun@hit.edu.cn}

\author[1]{\fnm{Kui} \sur{Jiang}}
\author[1]{\fnm{Xianming} \sur{Liu}}

\affil[1]{\orgdiv{Faculty of Computing}, \orgname{Harbin Institute of Technology}, \orgaddress{\city{Harbin} \postcode{150001},  \country{China}}}


\abstract{
Deep models have achieved significant process on single image super-resolution (SISR) tasks, in particular large models with large kernel ($3\times3$ or more). However, the heavy computational footprint of such models prevents their deployment in real-time, resource-constrained environments. Conversely, $1\times1$ convolutions bring substantial computational efficiency, but struggle with aggregating local spatial representations, an essential capability to SISR models. In response to this dichotomy, we propose to harmonize the merits of both $3\times3$ and $1\times1$ kernels, and exploit a great potential for lightweight SISR tasks. Specifically, we propose a simple yet effective fully $1\times1$ convolutional network, named Shift-Conv-based Network (SCNet). By incorporating a parameter-free spatial-shift operation, it equips the fully $1\times1$ convolutional network with powerful representation capability while impressive computational efficiency. Extensive experiments demonstrate that SCNets, despite its fully $1\times1$ convolutional structure, consistently matches or even surpasses the performance of existing lightweight SR models that employ regular convolutions. The code and pre-trained models can be found at \burl{https://github.com/Aitical/SCNet}.
}




\maketitle

\section{Introduction}
Single image super-resolution (SISR) aims at reconstructing a high-resolution (HR) image from its corresponding degraded low-resolution (LR) one. It has witnessed substantial advancements and gained more of the spotlight in research communities with the rapid development of deep learning\cite{survey_dnn_MIR,lightweight_SISR_survey_information_fusion}. The pioneering work SRCNN \cite{SRCNN} proposes to learn the mapping from LR inputs to HR ones by a convolutional neural network (CNN) and outperforms traditional approaches. Subsequently, many CNN-based work explore more effective architectures \cite{SRGAN,EDSR,RDN,DBPN}. Besides CNN architectures, a transformer-based architecture \cite{SwinIR} has been proposed and achieved state-of-the-art (SOTA) performance. 

However, the models mentioned above improve the SISR performance with very deep or complicated network architectures, leading to a heavy burden on parameter amounts and computational cost. This makes it difficult to deploy them in resource-constrained environments, such as mobile or edge devices. Consequently, there is a high demand for efficient and lightweight SR models. Many work have been proposed to reduce the amounts of parameters or floating-point operations (FLOPs) to achieve lightweight neural networks for SISR \cite{FSRCNN,CARN,SMSR,ECBSR,FDIWN,li2023fsr}.

The $3\times3$ convolution operation is the most widely used operation in CNN-based models due to its advantageous in balancing the model capacity and computational cost. While a larger kernel can promote better performance, it comes at the cost of a rapid increase in the number of parameters and computational cost \cite{convnet,largekernel}. Conversely, a smaller kernel with a size of $1\times1$ can reduce the number of parameters but impairs the learning ability because of the fixed receptive field and the absence of local feature aggregation with neighboring pixels. This leads us to the natural question: \textit{Can we achieve the best of both worlds and build a lightweight yet effective SR model with fully $1\times1$ convolutions?}

When directly replacing $3\times3$ convolution with $1\times1$ convolution, fixed receptive fields and the absence of local feature aggregation impair the model. To address this issue, we propose a novel method in this paper by extending the $1\times1$ convolution via the spatial-shift. 
It is worth noting that the spatial-shift operation is non-parametric, requiring no additional FLOPs, making it advantageous for highly optimized real-world applications \cite{Shift-Video,Sparse-Shift}. In detail, we divide the input feature map into different groups along the channel dimension and then apply the spatial-shift operation to each group with different spatial directions. It ensures that each pixel in the resulting feature map is assembled around features along the channel dimension, bridging the gap of representation capability to the $3\times3$ convolution, as shown in \cref{fig:spatial_shift_demo}.
We refer to this extended $1\times1$ convolution with local feature aggregation via the spatial-shift operation as the Shift-Conv layer (or SC layer for simplicity). Compared to the normal $3\times3$ convolution, the SC layer significantly reduces the number of parameters while maintaining comparable performance.

\begin{figure}
 \centering
 \includegraphics[width=0.75\textwidth]{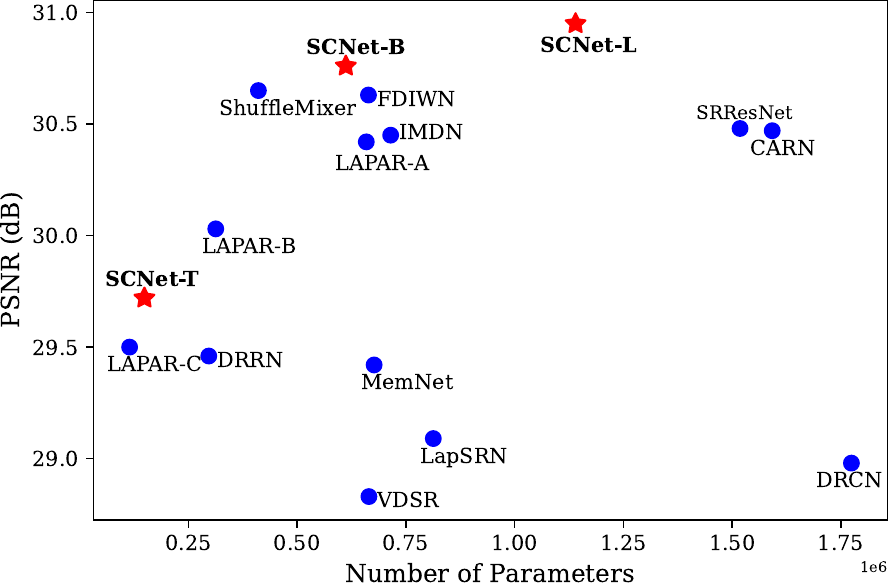}
 \caption{\textbf{PSNR} vs. \textbf{Parameters}. Comparisons with most recent efficient SISR models on Manga109 ($\times4$) test dataset.\label{fig:manga_psnr_ssim}}
 %-10pt
\end{figure}

Therefore, this paper proposes a lightweight yet effective SR model with fully $1\times1$ convolutional layers, containing extremely few parameters. The stride and direction hyper-parameters in the SC layer can be analogous to those in the normal $3\times3$ convolution when we set the stride as 1 in around eight directions. It is worth noting that different spatial priors can be achieved by selecting adaptive locations (even acting like deformable convolution \cite{deformable_conv}). The flexibility of different spatial priors enables the SC layer to reduce parameters while extending the receptive fields of the normal $3\times3$ convolution.
Following the widely used residual block \cite{EDSR}, we propose a shift-conv residual block, simplified as the SC-ResBlock. Furthermore, we propose a lightweight network, stacked by several SC-ResBlocks, named \textbf{SCNet}. The proposed SCNet is scalable to different model sizes and provides more opportunities to exploit wider or deeper architectures due to the few parameter amounts in the SC layer. We introduce three SCNets with different model sizes: tiny (T), base (B), and large (L), respectively. Moreover, the proposed SCNet is flexible to interpolate with extensive modules, such as widely used attention mechanisms, providing great potential for further study.
The performance of the proposed SCNets on the Manga109 test dataset ($\times4$) compared to other models of different sizes is shown in \cref{fig:manga_psnr_ssim}. The results demonstrate that the proposed SCNet achieves a better trade-off between SR results and the number of parameters.

Before diving into details, we summarize the main contributions of our work:
Firstly, we present the first fully $1\times1$ convolution-based SISR deep networks, shedding new light on the design of lightweight architectures.
Secondly, we investigate the feature aggregation in normal $3\times3$ convolution and extend $1\times1$ convolution with local feature aggregation by a manual spatial-shift operation against the channel dimension.
Lastly, we present extensive experimental results that verify the superiority of the proposed SCNet, along with detailed ablation studies that help understand the impact of various components and the scalability of the proposed SCNet. 

In the following section, we will first give some related work of lightweight image super-resolution methods in \cref{sec:related_work}. In \cref{sec:method}, we introduce and explain our proposed SCNet in detail. Then, \cref{sec:experiments} describes our training settings and experimental results, where we compare the performance of our approach to other state-of-the-art methods. Furthermore, though ablation studies are conducted to analyze the impact of different components in SCNet and the scalability of it. Finally, some conclusions are drawn in \cref{sec:conclusion}.

\section{Related Work \label{sec:related_work}}
Recently, deep learning methods have achieved dramatic improvements in SISR tasks \cite{survey,21survey}. Especially for CNN-based models, various well-designed CNN architectures explore to further improve the SISR performance \cite{VDSR,DRRN,EDSR}. Besides, attention mechanism like the channel attention \cite{channel_attention} has been introduced to SISR task as well \cite{RCAN,SAN,HAN}. Most recently, vision transformers have attracted great attention \cite{ViT,SwinT} and many work have been proposed to explore transformer-based architectures that achieve SOTA performance \cite{IPT,SwinIR, zhang2023practical}. In addition to encompassing architectures, some effort has been made to leveraging the SISR task with more learning patterns, such as neural network pruning \cite{network_pruning_SISR_pami}, contrastive learning \cite{PCL,model_contrastive_ir}, and knowledge distillation \cite{Knowledge-distillation-sisr}. Zhao \textit{et al.} \cite{loss_in_IR} embarked on an empirical examination of suitable objective functions. Wu \textit{et al.}\cite{PCL} innovated the contrastive learning framework for low-level SR tasks, providing an additional boost to the performance of existing methodologies. These diversified approaches to improving SISR continues to fuel the progression of this complex field.

In contrast to achieving advancing performance with a rapidly increased number of parameters and computational cost, many lightweight SISR models have been exploited by reducing parameters, especially for resource-limited devices \cite{CARN,IDN,IMDN,LAPAR,SMSR,ECBSR,FDIWN}. Hui \textit{et al.} proposed a deep information distillation network (IDN) \cite{IDN} and extended it into the information multi-distillation network (IMDN) \cite{IMDN}. Zhang \textit{et al.} \cite{ECBSR} proposed a real-time inference SR network by the re-parameterization strategy. Li \textit{et al.}\cite{sLWSR} proposed a super lightweight model with low computational complexity, named s-LWSR, by using a symmetric architecture, compression modules, and reduced activations. They commonly leverage the normal $3\times3$ convolutions and try to develop well-designed blocks to promote the performance.

In the last year, several work investigated some modern CNN-based architectures \cite{convnet,largekernel}. Liu \textit{et al.} explored a modern CNN-based architecture and introduced larger kernels that utilize $7\times7$ kernel size. Ding \textit{et al.} further brought the kernel size up to 31. Larger kernels bring larger receptive fields that significantly improve the capabilities of CNN-based networks compared to normal $3\times3$ convolution. Most recently, Liu \textit{et al.} \cite{shufflemixer} exploited the large kernel in the lightweight SR network, which utilizes the channel shuffle operation to further reduce the number of learnable features. 

Spatial-shift operation is widely adopted in various computer vision tasks. Several existing works, such as \cite{Shift_Shift,active_shift,Sparse-Shift}, have explored the use of spatial-shift operation in high-level tasks. Wu \textit{et al.} \cite{Shift_Shift} were the first to introduce the shift operation in convolution and proposed a compacted CNN model. Subsequently, adaptive and sparse shift operations were proposed in \cite{active_shift,Sparse-Shift}. Additionally, Lin \textit{et al.} \cite{Shift-Video} introduced the shift operation for temporal feature aggregation in videos. In the field of image super-resolution, Zhang \textit{et al.} \cite{ELAN} introduced the Efficient Long-range Attention Network (ELAN), incorporating a spatial-shift operation in its feed-forward network to enhance local feature aggregation. Our work, however, stands apart by fundamentally reimagining the network architecture with fully 1$\times$1 convolutions. Unlike existing methods that incorporate the spatial-shift operation as a minor component, our approach redefines the basic network architecture. This novel design emphasizes simplicity and efficiency, making a distinct contribution to the domain of super-resolution imaging.
 
In this paper, we focus on exploring an effective convolutional model for lightweight SISR tasks, specifically by converting $3\times3$ convolution-based models into fully $1\times1$ convolutional models. However, $1\times1$ convolution lacks local feature aggregation and is unable to learn effectively. To address this challenge, we propose an effective yet efficient SCNet, which employs a basic group shift strategy for local feature aggregation. In addition, we provide detailed benchmark comparisons and ablation studies, demonstrating the potential of SCNet for developing efficient SISR models. We believe that our work will contribute to the development of efficient SISR models for the research community. 

\begin{figure*}[t]
  \centering
 \includegraphics[width=\textwidth]{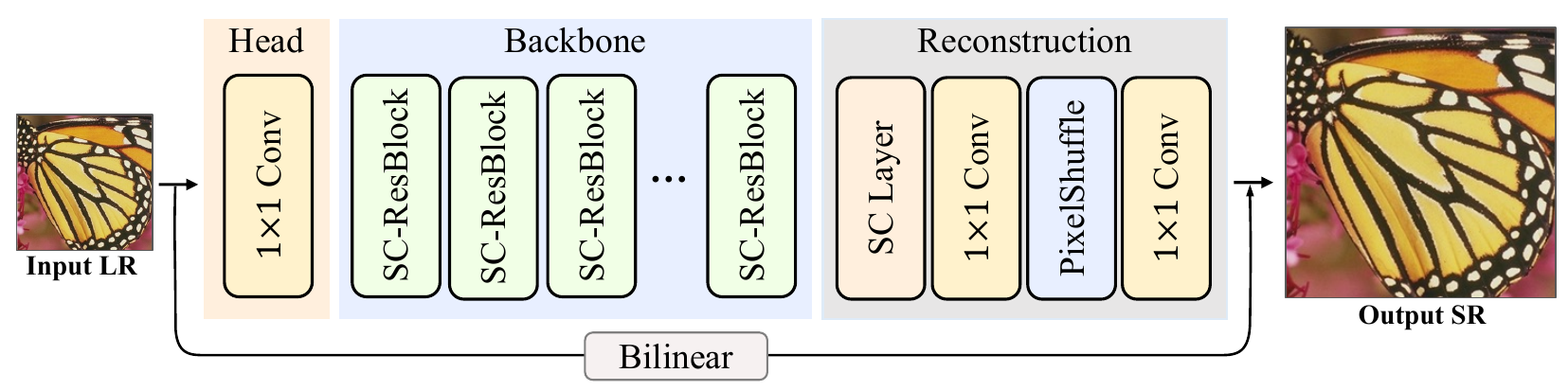} % framework_new
 \caption{The architecture of the proposed SCNet which is simply stacked by numerous basic residual blocks.}
 %-5.pt
\label{fig:framework}
\end{figure*}

\section{Methods \label{sec:method}}
In this section, we provide a detailed description of our proposed SCNet. We begin by introducing the general framework for SISR tasks. Subsequently, we present the implementation details of the different components in SCNet.

\subsection{Overview Architecture}
As shown in \cref{fig:framework}, numerous basic SR-ResBlocks stack the main backbone of the proposed SCNet followed by up-scaling layers to reconstruct high-resolution (HR) results.

Given the LR image $I^{LR}\in \mathbb{R}^{C \times H \times W}$where $H$, $W$, and $C$ are image height, width, and channel number, respectively. Firstly, a normal $1\times1$ convolution is utilized as the shallow feature extractor to map image space to a latent space. The shallow extractor is noted as ${N}_{head}$ and latent feature is $f_{head}=N_{head}(I^{LR})\in \mathbb{R}^{C_{latent}\times H \times W}$ where $C_{latent}$ is the channel dimension of the latent space.

 Main backbone $N_{main}$ is stacked by numerous basic SC-ResBlocks that are implemented by the shift-conv and $1\times1$ convolutional layers replacing the $3\times3$ convolutional layers in the normal residual block \cite{EDSR}. Here the main backbone $N_{main}$ takes shallow features $f_{head}$ as input and extracts deep features $f_{main}=N_{main}(f_{head})$.

Then given the extracted deep feature $f_{main}$, the up-scaling module is utilized to reconstruct HR results. We take the SC layer, ReLU, 1$\times$1 convolution, and the pixel-shuffle operation to build up-scaling module $N_{rec}$, and a normal $1\times1$ convolution is utilized to map the up-scaled feature into the output with 3 channels. In addition, we add the up-scaled LR images by bilinear interpolation and the super-resolved output is $I^{SR} = N_{rec}(f_{main})+\rm Bilinear(I^{LR})$. Finally, the SR network is trained by minimizing $L_{1}$ loss.

\begin{figure}[t]
  \centering
 \includegraphics[width=0.67\textwidth]{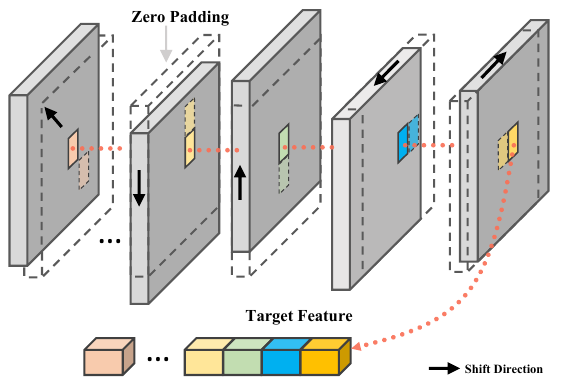} 
 \caption{Illustration of the spatial-shift operation, covering eight local regions. By rearranging the spatial positions of feature maps, spatial-shift operation enhances local spatial feature aggregation across channel groups without additional computational costs.}
 %-15.pt
\label{fig:spatial_shift_demo}
\end{figure}

\subsection{Shift-Conv Residual Block}

\textbf{Spatial-Shift Operation.}
Let us note the shift direction as $d\in \{1, 0, -1\}$, and take $d_{h}$ and $d_{w}$ for each side, respectively. Correspondingly, the strides are noted as $s_{h}$ and $s_{w}$. Then we can obtain the spatial-shift steps by combining direction and stride as $step=(d_{h}*s_{h},~d_{w}*s_{w})$, and the set of spatial-shift steps is $S=\{step_{i}, i=1,\dots,n\}$ where $n$ is the number of assembled features and $step_{i}$ presents the step for the $i$th local pixel-wise feature. If we want to take 8 local pixels around like the normal $3\times3$ convolution, the set of spatial-shift steps can be defined as $\{(0,~1),~(0,-1), ~(1,~0),~(1,~1),~(1,-1),~(-1,~0),~(-1,~1), ~(-1,-1)\}$. We utilize the $step_{i}$ to locate the target pixel feature and we can leverage pixels anywhere even with a long distance (just assign a large stride value). In addition, we can take different local aggregation schemes by setting different spatial-shift steps. For fair comparison and evaluating the effectiveness of the fully $1\times1$ convolutional SCNet, we take the local 8 pixels around like the normal $3\times3$ convolutional layer as the default.

\begin{algorithm}[htbp]
  \caption{PyTorch-style code for spatial-shift operation.}
  \label{alg:spatial_shift}
   
    \definecolor{codeblue}{rgb}{0.25,0.5,0.5}
    \definecolor{codekw}{rgb}{0.85, 0.18, 0.50}

    \lstset{
  backgroundcolor=\color{white},
  basicstyle=\fontsize{8.2pt}{8.2pt}\ttfamily\selectfont,
  columns=fullflexible,
  breaklines=true,
  captionpos=b,
  commentstyle=\fontsize{8.2pt}{8.2pt}\color{codeblue},
  keywordstyle=\fontsize{8.2pt}{8.2pt}\color{codekw},
}
\begin{lstlisting}[language=python]
# F: torch.nn.functional
def spatial_shift(f, steps, pad):
    """
    f [torch.Tensor]: input feature in (B, C, H, W)
    steps [Tuple(Tuple(int, int))]: parameters of the spatial-shift steps
    pad [int]: padding size
    """
    shift_groups = len(steps)
    B, C, H, W = f.shape
    group_dim = C//shift_groups
    f_pad = F.pad(f, pad)
    output = torch.zeros_like(f)
    
    for idx, step in enumerate(steps):
        s_h, s_w = step[0], step[1]
        output[:, idx*group_dim: (idx+1)*group_dim, :, :] = \
              f_pad[:, idx*group_dim:(idx+1)*group_dim, pad+s_h:pad+s_h+H, pad+s_w:pad+s_w+W]
    return output
\end{lstlisting}
\end{algorithm}

Given the input feature $f$, we uniformly split it into $n$ groups along the channel dimension where $n=|S|$, and $n$ thinner tensors $f^{i} \in \mathbb{R}^{\frac{C_{latent}}{n}\times H\times W}, i=1,\dots,\frac{C_{latent}}{n}$ are obtained. Then each separated feature group is shifted by the given step parameters and the shifted feature $f_{shift}$ is obtained. Each pixel feature in $f_{shift}$ contains local features around it along the channel dimension. Details of the spatial-shift operation are shown in \cref{fig:spatial_shift_demo}. Implementation of the spatial-shift operation is presented in \cref{alg:spatial_shift}. Here we adopt the vanilla Python implementation based on Pytorch for model training. Given the input feature $f$, it is separated and shifted with the hyper-parameter shift step, and we take the constant zero value for padding as the default.

\textbf{Shift-Conv Layer.} Since $1\times1$ convolutional operation works on the single pixel feature which impairs the modeling, here we explore the local feature aggregation explicitly by a simple spatial-shift operation that involves no parameters and FLOPs. The Shift-Conv layer (simplified as the SC layer) is stacked by a $1\times1$ convolutional layer and the spatial-shift operation, thus the SC layer extends the normal $1\times1$ convolution with local feature aggregation as well as fewer parameters.

\begin{figure}
 \centering
 \includegraphics[width=0.5\textwidth]{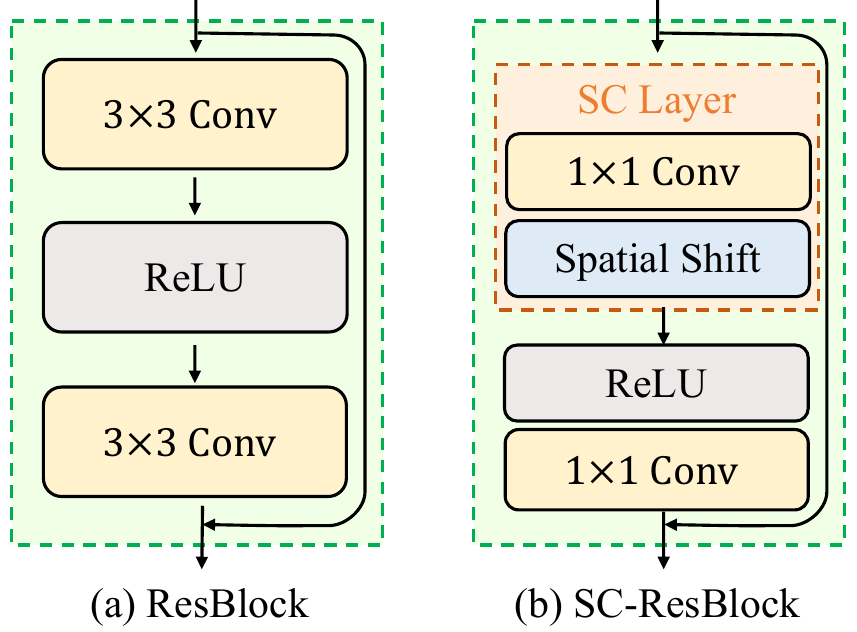}
     \caption{Side-by-side comparison of the basic ResBlock and our proposed SC-ResBlock. The proposed SC-ResBlock substantially reduces the complexity with fully $1\times1$ convolutions, while effectively aggregating local features by spatial-shift operation.}
\label{fig:scresblock}
\end{figure}

\textbf{Shift-Conv Residual Block.} 
As illustrated in \cref{fig:scresblock}{(a)}, the residual block proposed in \cite{EDSR} is widely used in SR networks. For a fair comparison, we modify and introduce the SC-ResBlock. As illustrated in \cref{fig:scresblock}(b), the SC-ResBlock contains the SC layer, ReLU, and a $1\times1$ convolution. Compared with the $3\times3$ convolution-based residual block, our SC-ResBlock significantly reduces the number of parameters and computational cost by adopting only $1\times1$ convolution.

\textbf{Remark.} Deep learning-based SISR techniques have made significant progress, but at the same time, their performance has become increasingly saturated. In this work, instead of exploring more complex network architectures, we look back to the minimal CNN unit and propose a lightweight SCNet, which employs fully $1\times1$ convolutions to reduce parameters and computational costs. 

The spatial-shift operation is not new in vision tasks and has been effectively applied for high-level vision tasks \cite{Shift_Shift,Sparse-Shift,Shift-Video}. It is worth noting that the goal of this work is not to present a novel operation algorithm. Instead, we attempt to build a benchmark SR network, which contains \textit{only} the simplest feature aggregation (spatial-shift operation) and the simplest feature extraction $(1 \times 1$ convolution). We hope this can shed some new light on the network design of low-level image restoration tasks, especially for lightweight architecture design.

\begin{table*}[!htb]
\renewcommand\arraystretch{1.4}
  \caption{Quantitative comparisons on five widely used benchmark datasets. The best and our results are highlighted in \underline{underline} and \textbf{bold} correspondingly. Avg. presents the average performance on test datasets besides Set5. \label{tab:main_results}}
  \centering
  
  \resizebox{0.975\textwidth}{!}{
  \begin{tabular}{cllccccccc}
    \toprule
   \multirow{2}{*}{Scale} &\multirow{2}{*}{Method} & \multirow{2}{*}{Avenue}  &\multirow{2}{*}{Params} & Set5 & Set14 &B100 & Urban100 & Manga109 & Avg. \\
    %\cmidrule(r){3-7}
       &  &  &    & PSNR/SSIM & PSNR/SSIM& PSNR/SSIM& PSNR/SSIM & PSNR/SSIM & PSNR/SSIM\\
    \midrule
  \multirow{18}{*}{$\times2$}
    % & SRCNN     & 57K     & 36.66/0.9542 & 32.45/0.9067 & 31.36/0.8879 & 29.50/0.8946 & 35.60/0.9663 \\
	%&	FSRCNN & 13K & 37.00/0.9558 &32.63/0.9088 &31.53/0.8920 &29.88/0.9020 &36.67/0.9710 \\

	&	LapSRN \cite{LapSRN} &CVPR'2017    & 251K    & 37.52/0.9591 & 32.99/0.9124 & 31.80/0.8952 & 30.41/0.9103 & 37.27/0.9740 & 33.12/0.9230 \\
	&	DRRN \cite{DRRN}  & CVPR'2017    & 298K    & 37.74/0.9591 & 33.23/0.9136 & 32.05/0.8973 & 31.23/0.9188 & 37.88/0.9749 & 33.60/0.9262 \\

	&	ECBSR-M10C32 \cite{ECBSR}  & ACM MM'2021 &	 95K &	 37.76/\underline{0.9609}&	 33.26/0.9146 &	 32.04/0.8986&	 31.25/0.9190 &	-/- & 32.18/0.9107 \\
	&   LAPAR-C \cite{LAPAR} &NeurIPS'2020  & 87K     & 37.65/0.9593 & 33.20/0.9141 & 31.95/0.8969 & 31.10/0.9178 & 37.75/0.9752 & 33.50/0.9260 \\
	&   LAPAR-B \cite{LAPAR} &NeurIPS'2020 &   250K &    \underline{37.87}/0.9600  &    \underline{33.39}/\underline{0.9162} &   \underline{32.10}/\underline{0.8987} &   \underline{31.62}/\underline{0.9235} &    38.27/\underline{0.9764} & \underline{33.85}/\underline{0.9287}\\
    &   \textbf{SCNet-T}& 2023 & 159K  & \textbf{37.85}/\textbf{0.9600} & \underline{\textbf{33.39}}/\textbf{0.9161} & \textbf{32.06}/\textbf{0.8981} & \textbf{31.50}/\textbf{0.9187} & \underline{\textbf{38.29}}/\underline{\textbf{0.9764}}
    & \textbf{33.81}/\textbf{0.9273} \\
    	\cmidrule(lr){2-10}
	&	VDSR \cite{VDSR}  & CVPR'2016         & 666K    & 37.53/0.9587 & 33.03/0.9124 & 31.90/0.8960 & 30.76/0.9140 & 37.22/0.9750 & 33.23/0.9244 \\

	&   CARN-M \cite{CARN} &ECCV'2018       & 412K    & 37.53/0.9583 & 33.26/0.9141 & 31.92/0.8960 & 31.23/0.9193 & 35.62/0.9420 & 33.01/0.9179\\
	&	IMDN \cite{IMDN}  &ACM MM'2019       & 694K    & 38.00/0.9605 & 33.63/0.9177 & 32.19/0.8996 & 32.17/0.9283 & 38.88/0.9774 &34.22/0.9308 \\
	&   LAPAR-A \cite{LAPAR} &NeurIPS'2020      & 548K    & 38.01/0.9605  & 33.62/0.9183 & 32.19/0.8999 & 32.10/0.9283 & 38.67/0.9772 & 34.15/0.9309\\

	&	FDIWN \cite{FDIWN} &AAAI'2022   & 629K & \underline{38.07}/\underline{0.9608} & \underline{33.75}/\underline{0.9201} & \underline{32.23}/\underline{0.9003} & \underline{32.40}/\underline{0.9305} &
 38.85/0.9774 & \underline{34.31}/\underline{0.9321} \\
	& ShuffleMixer \cite{shufflemixer} &NeurIPS'2022 & 394K & 38.01/0.9606 & 33.63/0.9180 & 32.17/0.8995 & 31.89/0.9257 & 38.83/0.9774 & 34.13/0.9302 \\
 
	&  \textbf{SCNet-B}& 2023 & 557K &	\underline{\textbf{38.07}}/\textbf{0.9607} & 	 \textbf{33.72}/\textbf{0.9188} & \underline{\textbf{32.23}}/\underline{\textbf{0.9003}}  & \textbf{32.24}/\textbf{0.9296} & \underline{\textbf{38.95}}/\underline{\textbf{0.9777}}
    & \textbf{34.29}/\textbf{0.9316} \\
	\cmidrule(lr){2-10}
	&	DRCN \cite{DRCN}& CVPR'2016        & 1,774K  & 37.63/0.9588 & 33.04/0.9118 & 31.85/0.8942 & 30.75/0.9133 & 37.55/0.9732 & 33.30/0.9231\\
	&	CARN \cite{CARN}  &ECCV'2018       & 1,592K  & 37.76/0.9590 & 33.52/0.9166 & 32.09/0.8978 & 31.92/0.9256 & 38.36/0.9765 & 33.97/0.9291 \\
	&	SRResNet \cite{SRGAN}   & CVPR'2017   & 1,370K  & 38.05/0.9607 & 33.64/0.9178 & 32.22/0.9002 &  32.23/0.9295 & 38.05/0.9607 & 34.04/0.9271\\
	% & SwinIR-light &ICCV'2021 &  878K & 38.14/0.9611   &33.86/0.9206 &  32.31/0.9012  & 32.76/0.9340 & 39.12/0.9783\\
		&   \textbf{SCNet-L} & 2023& 1,157K & \underline{\textbf{38.12}}/\underline{\textbf{0.9609}} & \underline{\textbf{33.90}}/\underline{\textbf{0.9206}} & \underline{\textbf{32.28}}/\underline{\textbf{0.9009}} & \underline{\textbf{32.46}}/\underline{\textbf{0.9315}} & \underline{\textbf{39.14}}/\underline{\textbf{0.9781}} &
  \underline{\textbf{34.45}}/\underline{\textbf{0.9328}}\\
     \midrule
     
	  \multirow{18}{*}{$\times3$} 

	&	DRRN \cite{DRRN} & CVPR'2017      & 298K    & 34.03/0.9244 & 29.96/0.8349 & 28.95/0.8004 & 27.53/0.8378 & 32.71/0.9379 & 29.79/0.8528\\
	&   LAPAR-C \cite{LAPAR} &NeurIPS'2020      & 99K     & 33.91/0.9235 & 30.02/0.8358 & 28.90/0.7998 & 27.42/0.8355 & 32.54/0.9373 & 29.72/0.8521 \\
    & LAPAR-B \cite{LAPAR}  &NeurIPS'2020    & 276K &  \underline{34.20}/\underline{0.9256} & \underline{30.17}/\underline{0.8387} & \underline{29.03}/\underline{0.8032}&  \underline{27.85}/\underline{0.8459} & \underline{33.15}/\underline{0.9417} &
    \underline{30.05}/\underline{0.8574}\\

	& \textbf{SCNet-T} &2023 & 147K  & \textbf{34.03}/\textbf{0.9244} & \textbf{29.99}/\textbf{0.8381} & \textbf{28.93}/\textbf{0.8017}& \textbf{27.65}/\textbf{0.8413} & \textbf{32.84}/\textbf{0.9403} &
    \textbf{29.85}/\textbf{0.8554}\\
    \cmidrule(lr){2-10}

	&	VDSR \cite{VDSR}    & CVPR'2016      & 666K    & 33.66/0.9213 & 29.77/0.8314 & 28.82/0.7976 & 27.14/0.8279 & 32.01/0.9340 & 29.44/0.8477\\
	&	LapSRN \cite{LapSRN} &CVPR'2017      & 502K    & 33.81/0.9220 & 29.79/0.8325 & 28.82/0.7980 & 27.07/0.8275 & 32.21/0.9350 & 29.47/0.8483\\
	&	IMDN \cite{IMDN} &ACM MM'2019         & 703K    & 34.36/0.9270 & 30.32/0.8417 & 29.09/0.8046 & 28.17/0.8519 & 33.61/0.9445 & 30.30/0.8607 \\
	&   LAPAR-A \cite{LAPAR}  &NeurIPS'2020      & 594K    & 34.36/0.9267 & 30.34/0.8421 & 29.11/0.8054 & 28.15/0.8523 & 33.51/0.9441 & 30.28/0.8610\\
	& LBNet \cite{LBNet} &IJCAI'2022   & 736K & 34.47/0.9277  & 30.38/0.8417  & 29.13/0.8061 &  \underline{28.42}/\underline{0.8559}  & 33.82/0.9460 & 30.44/0.8624\\
	& FDIWN \cite{FDIWN}  &AAAI'2022   & 645K & \underline{34.52}/\underline{0.9281} &  30.42/\underline{0.8438} & 29.14/\underline{0.8065} & 28.36/0.8567&  33.77/0.9456 & 30.42/\underline{0.8631}\\
	& ShuffleMixer \cite{shufflemixer} &NeurIPS'2022 &  415K &  34.40/0.9272 & 30.37/0.8423 & 29.12/0.8051 & 28.08/0.8498 & 33.69/0.9448 & 30.32/0.8605\\
   &\textbf{SCNet-B} &2023 & 589K & \textbf{34.44}/\textbf{0.9276} & \underline{\textbf{30.43}}/\textbf{0.8437} & \underline{\textbf{29.15}}/\textbf{0.8063} & \textbf{28.31}/{\textbf{0.8556}} & \underline{\textbf{33.86}}/\underline{\textbf{0.9462}} &
   \underline{\textbf{30.44}}/{\textbf{0.8630}}\\

   	\cmidrule(lr){2-10}
   	&	DRCN \cite{DRCN} & CVPR'2016   & 1,774K  & 33.82/0.9226 & 29.76/0.8311 & 28.80/0.7963 & 27.15/0.8276 & 32.24/0.9343 &  29.49/0.8473\\
	&	CARN \cite{CARN}&ECCV'2018       & 1,592K  & 34.29/0.9255 & 30.29/0.8407 & 29.06/0.8034 & 28.06/0.8493 & 33.50/0.9440 & 30.23/0.8594\\

	&	SRResNet \cite{SRGAN} & CVPR'2017   & 1,554K  & 34.41/0.9274 & 30.36/0.8427 & 29.11/0.8055 & 28.20/0.8535 & 33.54/0.9448 & 30.30/0.8616\\
    & SMSR \cite{SMSR}  &CVPR'2021  &  993K & 34.40/0.9270 & 30.33/0.8412&  29.10/0.8050 & 28.25/0.8536&  33.68/0.9445 & 30.34/0.8611\\
   & \textbf{SCNet-L} &2023 &1,107K & \underline{\textbf{34.53}}/\underline{\textbf{0.9284}} & \underline{\textbf{30.49}}/\underline{\textbf{0.8452}} & \underline{\textbf{29.20}}/\underline{\textbf{0.8076}} & \underline{\textbf{28.47}}/\underline{\textbf{0.8588}} & \underline{\textbf{34.08}}/\underline{\textbf{0.9475}} &
   \underline{\textbf{30.56}}/\underline{\textbf{0.8648}}\\
    \midrule

\multirow{18}{*}{$\times4$}

	&	DRRN \cite{DRRN} & CVPR'2017 & 297K  & 31.68/0.8888 & 28.21/0.7720 & 27.38/0.7284 & 25.44/0.7638 & 29.46/0.8960 &  27.62/0.7901\\
	&   ECBSR-M10C32 \cite{ECBSR}  &ACM MM'2021     & 98K   & 31.66/0.8911 & 28.15/0.7776 & 27.34/0.7363 & 25.41/0.7653 & -/- & 26.97/0.7597 \\
    &  $\text{s-LWSR}_{16}$ \cite{sLWSR}& TIP'2020 & 144K & 31.62/0.8860 &27.92/0.7700 & 27.35/0.7290 & 25.36/0.762 & -/-  & 26.87/0.7537\\
	&   LAPAR-C \cite{LAPAR}  &NeurIPS'2020         & 115K  & 31.72/0.8884 & 28.31/0.7740 & 27.40/0.7292 & 25.49/0.7651 & 29.50/0.8951 & 27.68/0.7909\\
    &   LAPAR-B \cite{LAPAR}  &NeurIPS'2020         & 313K & \underline{31.94}/\underline{0.8917} & \underline{28.46}/\underline{0.7784} & \underline{27.52}/\underline{0.7335} & \underline{25.85}/\underline{0.7772} &  \underline{30.03}/\underline{0.9025} &
    \underline{27.97}/\underline{0.7979}\\

   & \textbf{SCNet-T}  &  2023 & 149K    & \textbf{31.82}/\textbf{0.8904} & \textbf{28.36}/\textbf{0.7764} & \textbf{27.39}/\textbf{0.7309}    & \textbf{25.59}/\textbf{0.7696}  & \textbf{29.72}/\textbf{0.9000} &
  \textbf{27.77}/\textbf{0.7942}\\
		\cmidrule(lr){2-10}
  
	&	VDSR \cite{VDSR}  & CVPR'2016       & 665K       & 31.35/0.8838 & 28.01/0.7674 & 27.29/0.7251 & 25.18/0.7524 & 28.83/0.8809 & 27.33/0.7815\\

	& CARN-M \cite{CARN} &ECCV'2018 &  412K & 31.92/0.8903 & 28.42/0.7762 & 27.44/0.7304 & 25.62/0.7694  & 25.62/0.7694 & 26.78/0.7614\\
	& SRFBN-S \cite{SRFBN}& CVPR'2019 &  483K & 31.98/0.8923 & 28.45/0.7779 & 27.44/0.7313 & 25.71/0.7719 & 29.91/0.9008 & 27.88/0.7955\\
	&	IMDN \cite{IMDN} &ACM MM'2019       & 715K       & 32.21/0.8948 & 28.58/0.7811 & 27.56/0.7353 & 26.04/0.7838 & 30.45/0.9075 & 28.16/0.8019\\

& $\text{s-LWSR}_{32}$ \cite{sLWSR}& TIP'2020 & 571K & 32.04/0.8930 & 28.15/0.7760 & 27.52/0.734 & 25.87/0.7790 & -/- & 27.18/0.7630\\

 & LAPAR-A \cite{LAPAR} &NeurIPS'2020      & 659K       & 32.15/0.8944 &28.61/0.7818 &27.61/0.7366 &26.14/0.7871 &30.42/0.9074 & 28.20/0.8032\\
	& ECBSR-M16C64 \cite{ECBSR}   &ACM MM'2021 & 603K & 31.92/0.8946& 28.34/0.7817 &27.48/0.7393 &25.81/0.7773& -/- & 27.21/0.7661\\
	& LBNet \cite{LBNet}   &IJCAI'2022  & 742K &  \underline{32.29}/\underline{0.8960} &   28.68/0.7832 &   27.62/0.7382 &   26.27/0.7906  &  30.76/0.9111 & 28.33/0.8057\\
    & FDIWN \cite{FDIWN}   &AAAI'2022  & 664K & 32.23/0.8955& 28.66/0.7829 & 27.62/0.7380 & 26.28/0.7919 & 30.63/0.9098 & 28.29/0.8057\\
    & ShuffleMixer \cite{shufflemixer} &NeurIPS'2022  & 411K & 32.21/0.8953 &28.66/0.7827 &27.61/0.7366& 26.08/0.7835 &30.65/0.9093 & 28.25/0.8030\\
    
     & \textbf{SCNet-B}& 2023 & 578K & \textbf{32.26}/\textbf{0.8959}	& \underline{\textbf{28.70}}/\underline{\textbf{0.7844}} & \underline{\textbf{27.64}}/\underline{\textbf{0.7382}} & \underline{\textbf{26.28}}/\underline{\textbf{0.7917}} & \underline{\textbf{30.76}}/\underline{\textbf{0.9119}} &
     
     \underline{\textbf{28.35}}/\underline{\textbf{0.8066}}\\
     
    \cmidrule(lr){2-10}
    &	DRCN \cite{DRCN}  & CVPR'2016         & 1,774K     & 31.53/0.8854 & 28.02/0.7670 & 27.23/0.7233 & 25.14/0.7510 & 28.98/0.8816 & 27.34/0.7807\\
&	LapSRN \cite{LapSRN} &CVPR'2017  & 813K       & 31.54/0.8850 & 29.19/0.7720 & 27.32/0.7280 & 25.21/0.7560 & 29.09/0.8845 & 27.70/0.7851\\
	&	CARN \cite{CARN}  &ECCV'2018       & 1,592K     & 32.13/0.8937 & 28.60/0.7806 & 27.58/0.7349 & 26.07/0.7837 & 30.47/0.9084 & 28.18/0.8019\\
	&	SRResNet \cite{SRGAN}  & CVPR'2017   & 1,518K     & 32.17/0.8951 & 28.61/0.7823 & 27.59/0.7365 & 26.12/0.7871 & 30.48/0.9087 & 28.20/0.8036 \\
	& SMSR \cite{SMSR} &CVPR'2021  &  1,006K & 32.12/0.8932 & 28.55/0.7808 & 27.55/0.7351 & 26.11/0.7868 & 30.54/0.9085 & 28.19/0.8028 \\

   & \textbf{SCNet-L}   & 2023 &  1,140K  & \underline{\textbf{32.37}}/\underline{\textbf{0.8973}} & \underline{\textbf{28.79}}/\underline{\textbf{0.7861}} & \underline{\textbf{27.70}}/\underline{\textbf{0.7400}} & \textbf{26.44}/\underline{\textbf{0.7962}} & \underline{\textbf{30.95}}/\underline{\textbf{0.9137}} &
   \underline{\textbf{28.47}}/\underline{\textbf{0.8090}}\\
    \bottomrule
 \end{tabular}}
\end{table*}

\section{Experiments \label{sec:experiments}}
In this section we will describe the detailed evaluation experiments. Firstly, we introduce the experiment settings and comparison methods. Then quantitative and qualitative results are reported on some public datasets of SOTA light-weight methods and our proposed method. Moreover, we provide in-depth comparisons to evaluate the efficiency of the proposed SCNet with regard to the inference latency. Lastly, we provide though ablation studies to analyze the impact of different components especially for the Shift-Conv layer. Furthermore, we evaluate the scalability of SCNet by applying extensive modules to it.

\subsection{Experiment Setup}
\textbf{Training Settings.}
We crop the image patches with the fixed size of $64\times64$ for training, and the counterpart LR patches are downsampled by Bicubic interpolation. All the training patches are augmented by randomly horizontally flipping and rotation. We set the batch size to 32 and utilize the ADAM \cite{ADAM} optimizer with the settings of $\beta_1$ = 0.9, $\beta_2$ = 0.999. The initial learning rate is set as $2\times10^{-4}$.

\textbf{Datasets and Metrics.}
Following \cite{shufflemixer,LAPAR}, we take 800 images from DIV2K \cite{DIV2K} and 2650 images from Flickr2K for training. Datasets for testing include Set5 \cite{Set5}, Set14 \cite{Set14}, B100 \cite{B100}, Urban100 \cite{Urban100}, and Manga109 \cite{Manga109} with the up-scaling factor of 2, 3, and 4. For comparison, we measure Peak Signal-to-Noise Ratio (PSNR) and Structural Similarity Index Measure (SSIM) on the Y channel of transformed YCbCr space.

\textbf{Comparison methods.}
We compare the proposed SCNet with representative efficient SR models, including SRCNN \cite{SRCNN}, VDSR \cite{VDSR}, LapSRN \cite{LapSRN}, DRRN \cite{DRRN}, CARN \cite{CARN}, IMDN \cite{IMDN}, LAPAR \cite{LAPAR}, SMSR \cite{SMSR}, ECBSR \cite{ECBSR}, LBNet \cite{LBNet}, FDIWN \cite{FDIWN}, and ShuffleMixer \cite{shufflemixer} on $\times2$, $\times3$, and $\times4$ up-scaling tasks.

\begin{figure}[t]
  \centering
 \includegraphics[width=\textwidth]{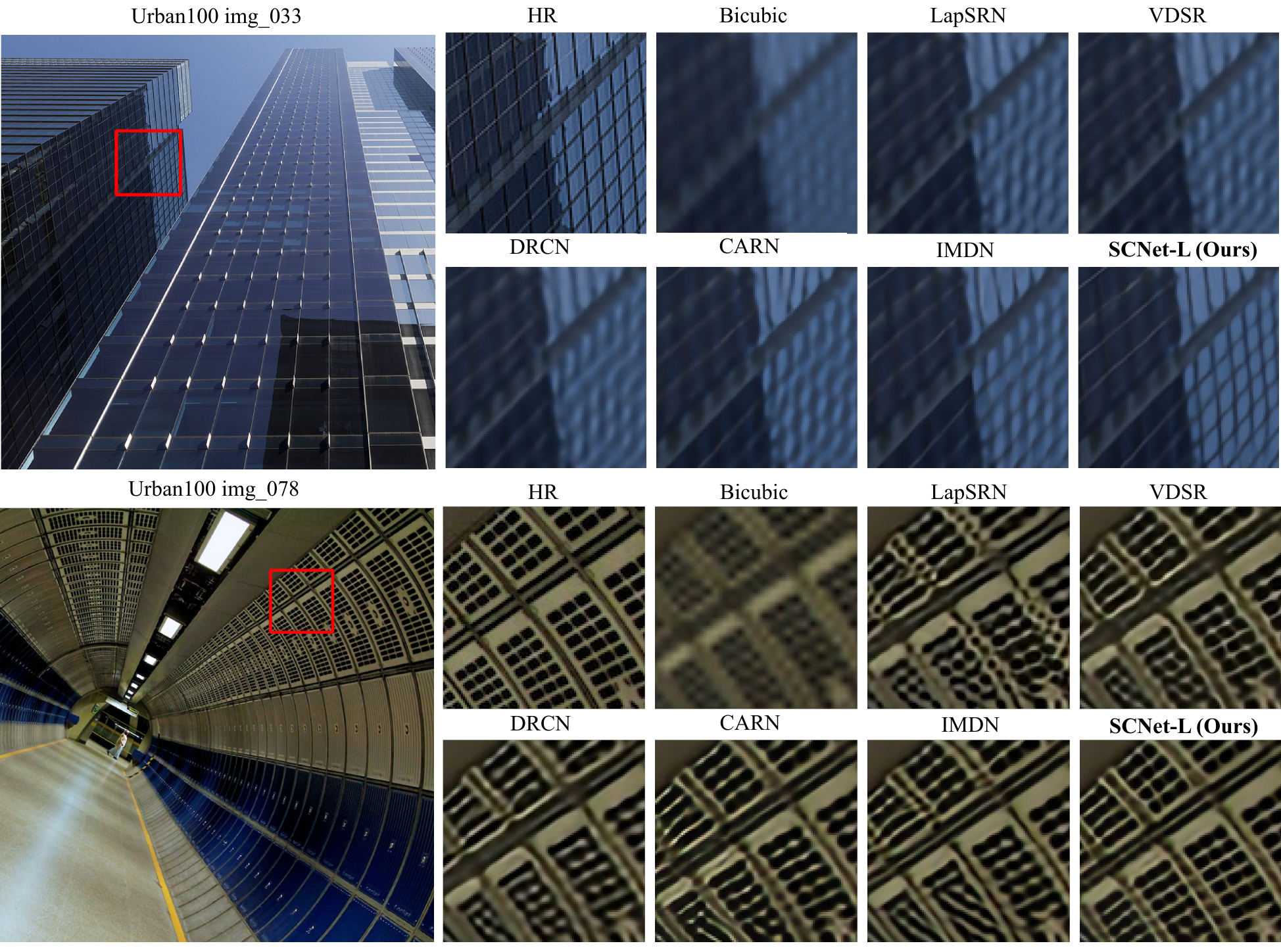}
 \caption{Visual comparisons on images with fine details on Urban100 test dataset (\textbf{Zoom in for more details}).}
\label{fig:SCNet-LPatch}
\end{figure}

\begin{figure}[ht]
  \centering
 \includegraphics[width=\textwidth]{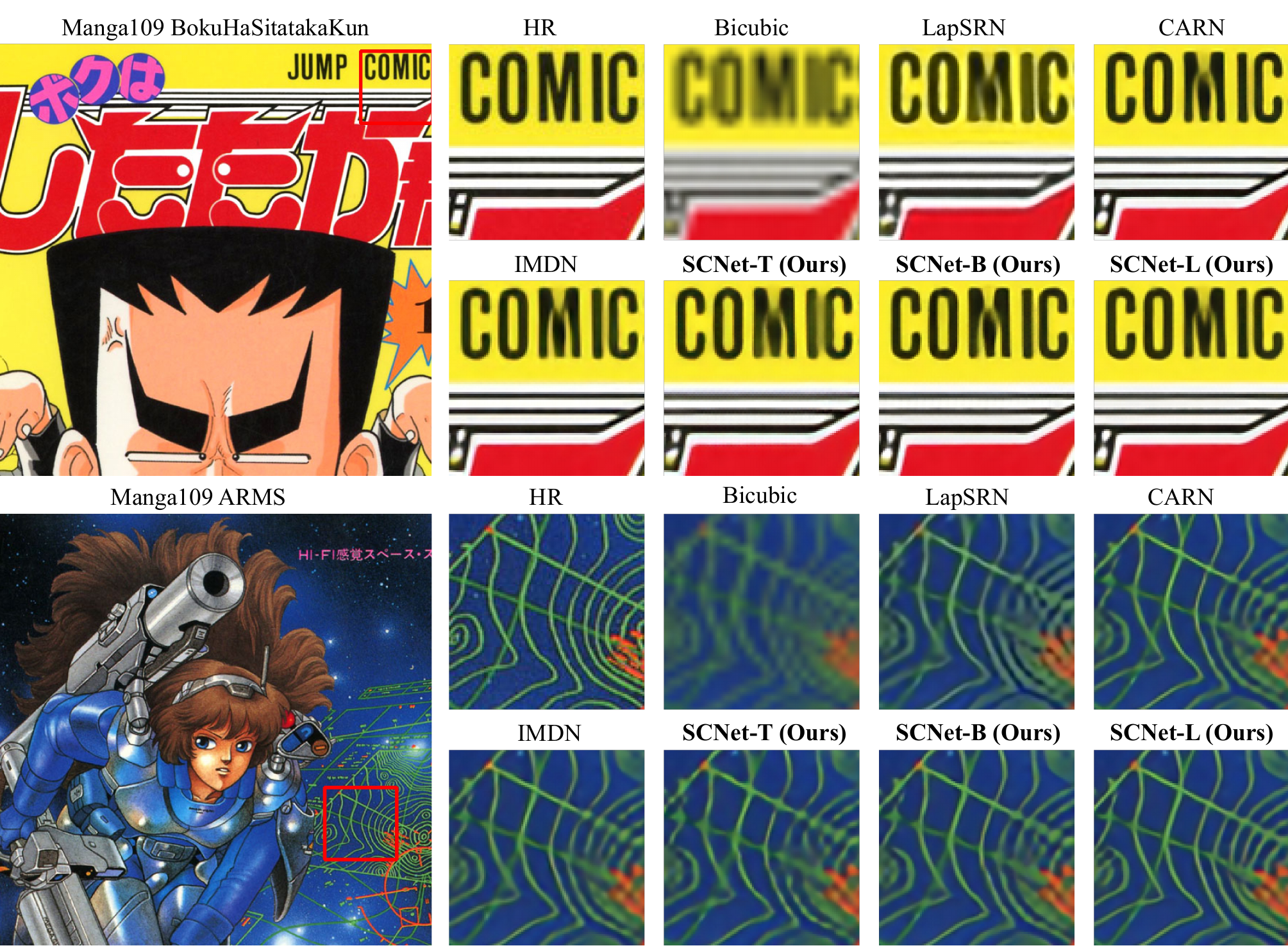}
 \caption{Visual comparisons on images with fine details on Manga109 test dataset (\textbf{Zoom in for more details}).}
\label{fig:SCNet-Manga109}
\end{figure}
\subsection{Main Results}

Benefiting from the extremely few parameters in SC layer, there are more opportunities for us to explore different architectures. In detail, simply stacked by the basic SC-ResBlock, we exploit three SCNets with differen model sizes that contain larger latent dimensions up to 128 channels and deeper architectures up to 64  blocks.

\textbf{Quantitative Evaluation.}
The performance of different SR models on five test datasets with scales 2, 3, and 4 is compared and reported in \cref{tab:main_results}. Along with PSNR and SSIM results, we also report the number of parameters. Besides LAPAR-B \cite{LAPAR}, our SCNet-T outperforms all the tiny models when the number of parameters is less than 400k, demonstrating its effectiveness. It is reasonable to note that LAPAR-B contains nearly twice as many parameters. When the number of parameters is between 400k and 800k, SCNet-B outperforms some larger models such as IMDN \cite{IMDN}, LAPAR-A \cite{LAPAR}, and FDIWN \cite{FDIWN} on all scales. Specifically, SCNet-B achieves advanced results on all test datasets besides Set5 compared to LBNet \cite{LBNet}, which contains well-designed architectures and more parameters. Furthermore, according to the average performance in \cref{tab:main_results}, one can observe that SCNet-B matches or even outperforms existing models across all scales, particularly for the x4 SR task. This effectively demonstrates the capability of our SCNet, which solely relies on $1\times1$ convolutions, to adeptly handle local feature aggregation for SR tasks. Lastly, the proposed SCNet-L outperforms DRCN \cite{DRCN}, CARN \cite{CARN}, SMSR \cite{SMSR}, and SRResNet \cite{SRGAN} and obtains the new SOAT performance in all test cases. SCNet-L achieves remarkable gains \textbf{0.26}/\textbf{0.0047} and \textbf{0.28}/\textbf{0.0062} in the terms of PSNR and SSIM compared to IMDN and SRResNet, respectively, demonstrating its effectiveness and scalability. Benefiting from the extremely few parameters in SC layer, there are more opportunities for us to explore different architectures. In detail, simply stacked by the basic SC-ResBlock, we exploit three SCNets with different model sizes that exploit larger latent dimensions up to 128 and deeper layers up to 64. We posit that by examining the results across varying architectures, we can provide a deeper understanding of the proposed SCNet and its performance nuances.

\begin{table}[tp]
  \caption{Complexity comparisons. The FLOPs is measured with the fixed $256\times 256$ LR input for scale 4.  * presents the average of test datasets besides Set5.\label{tab:psnr_vs_flops}}
  \centering
  \resizebox{0.7\textwidth}{!}{
  \begin{tabular}{cccc}
    \toprule

 Method & Avg. PSNR$^{*}$ (dB) & Params (K) & FLOPs (G)  \\
    \midrule
LAPAR-C   & 27.68 & 115 & 34 \\
\textbf{SCNet-T}  & 27.77 & 149 & 20 \\
\midrule
LAPAR-A   & 28.20 & 659 & 112 \\
ShuffleMixer & 28.25 & 411 & 32 \\
{ELAN} & {28.48} & {601} & {58} \\
\textbf{SCNet-B}   & 28.35 & 578 & 46 \\

\midrule
SRResNet   & 28.20 & 1,518 & 166 \\
\textbf{SCNet-L}   & 28.47 & 1,140 & 113 \\
\bottomrule
  \end{tabular}}
\end{table}

\textbf{{Efficiency of SCNet.}} In addition, we also report computational comparisons in \cref{tab:psnr_vs_flops} and show that SCNets obtain the advancded trade-off between performance, parameter count, and FLOPs compared to the LAPAR \cite{LAPAR}, ShuffleMixer \cite{shufflemixer}, SRResNet \cite{SRGAN}, and Transformer-based ELAN \cite{ELAN}. While ShuffleMixer demonstrates lower complexity, our SCNets leverage a simple yet effective residual architecture, establishing a new benchmark in lightweight super-resolution. This work deliberately focuses on the foundational aspects of SR architecture; exploring more intricate operations is reserved for future endeavors. Notably, SCNet showcases a significant improvement over CNN-based SRResNet, underscoring the effectiveness of our Shift-Conv layer. Additionally, our results reveal a notable performance gap between CNN-based methods and Transformer-based ELAN. However, SCNet serves as a significant bridge, narrowing this gap and demonstrating a promising fusion of simplicity and efficiency.

Shift operation is promising for designing lightweight models as they require no extra computational cost. For our proposed SCNet, which contains fully $1\times1$ convolutions, we find that the $1\times1$ convolution and spatial-shift operation in the SC layer can be fused as one optimal operation by re-indexing output values of the matrix dot product according to the shift step. To evaluate this fusion, we adopt the widely used C++ inference library NCNN, and the results are presented in \cref{tab:inference}. All models were converted from their official release without additional optimization. Compared to existing models, ShuffleMixer needs to be highly optimized for deployment due to complicated operations such as LayerNorm, channel-split-shuffle, and depth-wise convolution. IMDN and SRResNet perform well due to highly optimized implementations for widely used $3\times3$ convolutions. Finally, the proposed SCNet with vanilla fused Shift-Conv obtains comparable performance.

Notably, SCNet contains only one type of computational operation ($1\times1$ convolution). This simplicity makes it friendly and practical to achieve optimized implementation, which we believe will make it suitable for real-world applications in the future.

\begin{table}[tbp]
\begin{center}
\caption{Inference time comparison with $256\times256$ LR input. } \label{tab:inference}

\resizebox{0.75\textwidth}{!}{
  \begin{tabular}{lccccc}
    \toprule
Method & IMDN & ShuffleMixer & \textbf{SCNet-B} &\textbf{SCNet-L} & SRResNet \\
\midrule
Latency (ms) & 172 & 499 &\textbf{162} & \textbf{208} & 222 \\
\bottomrule
\end{tabular}

} 
\end{center}
\end{table}

In general, SCNets with all $1\times1$ convolutions obtain comparable and sometimes even better results than SR models with normal $3\times3$ convolutions with a larger model size, demonstrating the effectiveness of the proposed SCNets. In this regard, we believe that there are more opportunities to exploit efficient architectures for lightweight image restoration based on the proposed SCNet.

\begin{table*}[tb]
\renewcommand\arraystretch{1.2}
  \caption{Results of different selected positions. Based on the same SCNet architecture, containing 16 SC-ResBlocks with 128 channel dimensions, we replace the default Shift8 step with different settings as shown in \cref{fig:local_selection}.}\label{tab:appendix_shift_step}
  %\vskip -5pt
  \centering
  \resizebox{\textwidth}{!}{
  \begin{tabular}{clccccccc}
\toprule
  \multirow{2}{*}{Scale}&\multirow{2}{*}{Shift Step}    &\multirow{2}{*}{Params} & \multirow{2}{*}{FLOPs} & Set5 & Set14 &B100 & Urban100 & Manga109 \\
    %\cmidrule(r){3-7}
      &  &     &   & PSNR/SSIM & PSNR/SSIM& PSNR/SSIM& PSNR/SSIM& PSNR/SSIM \\
    \midrule
\multirow{4}{*}{$\times4$}
& Shift4-Cross       & 612K  & 78G  & 32.14/0.8946 & 28.61/0.7819 & 27.58/0.7360 & 26.05/0.7836 & 30.48/0.9086 \\
& Shift4-Diag    & 612K & 78G  & 31.83/0.8898 & 28.39/0.7769 & 27.44/0.7314 & 25.65/0.7705 & 29.90/0.9015 \\
& Shift8          & 612K & 78G  & 32.16/0.8949 & 28.65/0.7830 & 27.60/0.7368 & 26.16/0.7864 & 30.58/0.9100 \\
& Shift8-Dilated  & 612K  & 78G & 32.19/0.8953 & 28.67/0.7832 & 27.60/0.7369 & 26.14/0.7868  & 30.61/0.9102 \\
& Shift16         & 612K & 78G  & 32.10/0.8941 & 28.57/0.7812 & 27.55/0.7355 & 26.02/0.7833 & 30.34/0.9075 \\
\bottomrule
  \end{tabular}   
  }
\end{table*}

\textbf{Qualitative Evaluation.}
We conducted a visual quality comparison of SR results between our proposed SCNet-L and five representative models, including LapSRN \cite{LapSRN}, VDSR \cite{VDSR}, DRCN \cite{DRCN}, CARN \cite{CARN}, and IMDN \cite{IMDN}, for up-scaling tasks of $\times2$, $\times3$, and $\times4$. The $\times4$ SR results on Urban100 test dataset are presented in \cref{fig:SCNet-LPatch}. One can find that the results of CARN and IMDN appear blurry and contain more artifacts compared to our SCNet-L, which is able to recover the main structures with clear and sharp textures. In addition, results of the proposed SCNet with different model capacity are presented in \cref{fig:SCNet-Manga109}. When we have a look at image `BokuHaSitatakaKu`, we can find that even SCNet-B can achieve clearer characters compared to results of IMDN or CARN.

\begin{figure}[tb]
 \centering
 \includegraphics[width=\textwidth]{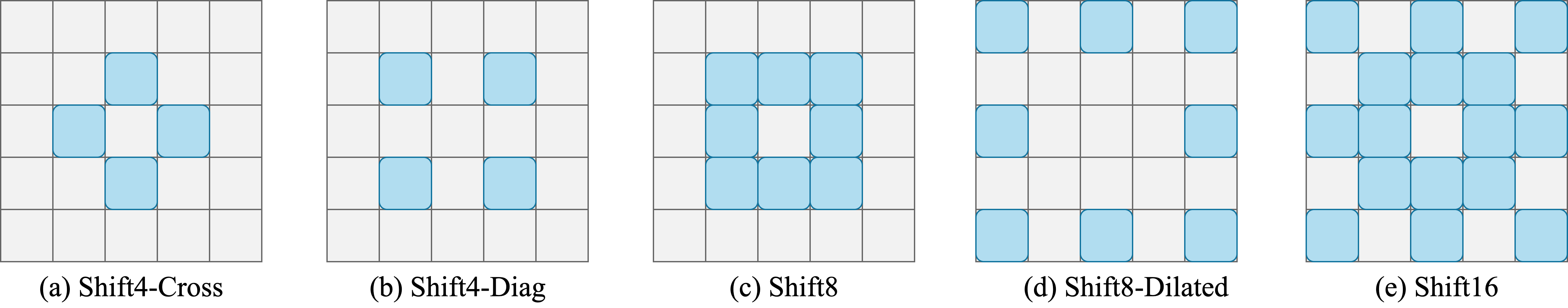}
 \caption{Illustration of different spatial-shift steps. We provide five different feature aggregation patterns to analyze its impact to model capacity.}
\label{fig:local_selection}
%-5.pt
\end{figure}

\subsection{Ablation Analysis}
The core contribution in this paper is to propose a fully $1\times1$ convolutional network for SISR. To better understand the impact of different components of our SCNet, comprehensive ablation studies are presented in this section.

\begin{figure}[tb]
 \centering
 \includegraphics[width=0.6\textwidth]{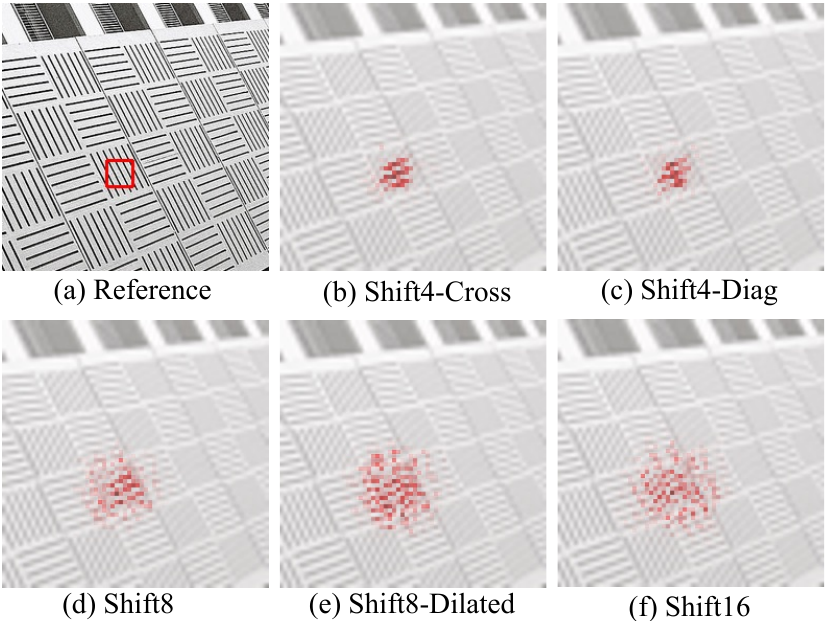} %/step_lam.png %
 \caption{LAM \cite{LAM} comparisons between different shift steps. Each shift step configuration results in varied feature aggregation patterns and is crucial to receptive fields. \label{fig:appendix_shift_step_lam}}
\end{figure}

\textbf{The Impact of Steps in SC Layer.}
Compared to the normal $3\times3$ convolution, $1\times1$ convolution lacks spatial feature aggregation. To address this, we introduce the spatial-shift operation to aggregate local features. The hyper-parameter shift step, which determines the aggregated local pixels, plays a key role in this operation. To better understand the impact of the shift step, we adopt our basic model, SCNet with 16 SC-ResBlocks and 128 channel dimensions, and re-train it with five different shift step settings as shown in \cref{fig:local_selection}. The first and second patterns involve four local positions from the horizontal and vertical directions (Shift4-Cross) and diagonal directions (Shift4-Diag), respectively. The remaining patterns are dense 8 pixels around (Shift8), dilated 8 pixels (Shift8-Dilated), and 16 pixels that combine Shift8 and Shift8-Dilated (Shift16). We report the results in \cref{tab:appendix_shift_step}. We utilize LAM \cite{LAM} to visualize the receptive fields of different spatial steps, as shown in \cref{fig:appendix_shift_step_lam}. In general, we observe that local feature aggregation is critical in the following three aspects.

\emph{Neighboring Feature Aggregation.}
The models with Shift4-Cross and Shift4-Diag are inferior to the model with default Shift8, indicating that feature aggregation patterns in Shift4-Cross and Shift4-Diag complement each other and the aggregation of neighboring pixels, like the normal $3\times3$ convolution, is essential for SR network. We observe that the Shift4-Diag can enable successful learning in the SR network, but it results in the worst performance, likely due to the loss of information during the spatial-shift operation. As we use a constant value of 0 for padding, the diagonal shift removes twice the number of pixels on two sides compared to Shift4-Cross.

%calculated with $256\times256$ LR input.

\begin{table*}[ht]
\renewcommand\arraystretch{1.2}
  \caption{Results of SCNets with different capacity. The number of the SC-Resblock and latent dimension  are simplified as the B and D.  \label{tab:appendix_capacity}}
  %5.pt
  \centering
  \resizebox{\textwidth}{!}{
  \begin{tabular}{clccccccc}
    \toprule
  \multirow{2}{*}{Scale}&\multirow{2}{*}{Model Size}    &\multirow{2}{*}{Params} &\multirow{2}{*}{FLOPs} &Set5 & Set14 &B100 & Urban100 & Manga109 \\
       &  &  &    & PSNR/SSIM & PSNR/SSIM& PSNR/SSIM& PSNR/SSIM& PSNR/SSIM \\
    \midrule
\multirow{4}{*}{$\times4$}
& B16D64  & 149K & 20G  & 31.82/0.8904 & 28.36/0.7764 & 27.39/0.7309 & 25.59/0.7696 & 29.72/0.9000 \\
& B32D64  & 312K & 29G  & 32.08/0.8939 & 28.59/0.7816 & 27.57/0.7357 & 26.01/0.7829 & 30.42/0.9079 \\
& B64D64  & 578K  & 46G & 32.26/0.8959 & 28.70/0.7844 & 27.64/0.7382 & 26.28/0.7917 & 30.76/0.9119 \\
& B16D128 & 612K & 78G  & 32.16/0.8949 & 28.65/0.7830 & 27.60/0.7368 & 26.16/0.7864 & 30.58/0.9100 \\
& B32D128 & 1,140K & 113G &32.37/0.8973 & 28.79/0.7861 & 27.70/0.7400 & 26.44/0.7962 & 30.95/0.9137 \\
\bottomrule
  \end{tabular}}
  %5.pt
\end{table*}

\begin{figure*}[!htbp]
 \centering
 \includegraphics[width=\textwidth]{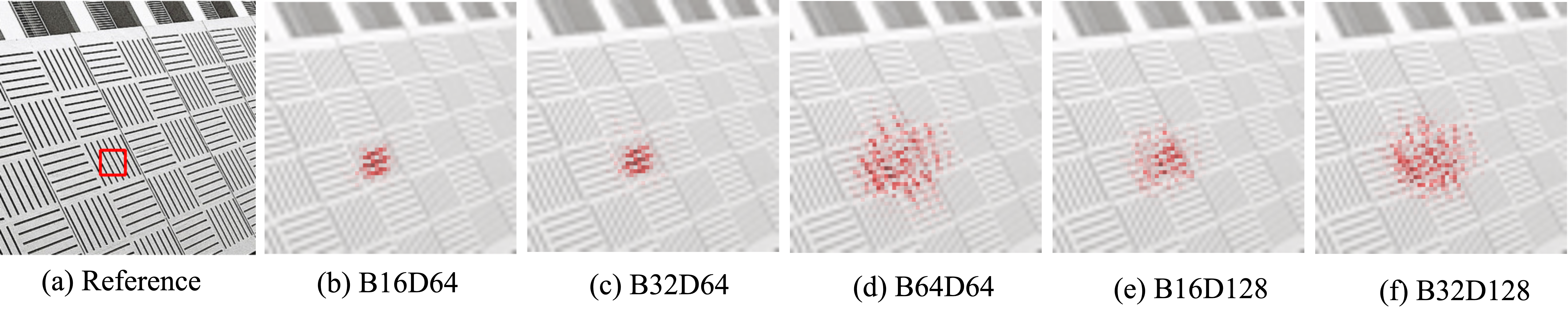}
 \caption{LAM \cite{LAM} comparisons between different architectures of SCNet.}
\label{fig:appendix_arch_lam}
\end{figure*}

\emph{Receptive Field.}
Based on the default Shift8 step, we extend it to Shift8-Dilated, as shown in \cref{fig:local_selection}(d). The dilated SCNet obtains slightly better performance than the default except for Urban100. According to \cref{fig:appendix_shift_step_lam}, a larger receptive field can be obtained by Shift8-Dilated, demonstrating that different feature aggregation patterns can be obtained through spatial-shift steps, like the normal dilated convolution.

\emph{Group Dimension.}
 Additionally, we combine the default Shift8 with Shift8-Dilated to obtain Shift16, shown in \cref{fig:local_selection}(e). Compared to Shift8 and Shift8-Dilated, SCNet with Shift16 obtains an even larger receptive field but has worse performance, as summarized in \cref{fig:appendix_shift_step_lam} and \cref{tab:appendix_shift_step}. We attribute this to the reduced feature dimensions of each shift group, which hampers feature extraction. Since the dimension of the latent feature is fixed, the number of shift group dimensions in Shift16 is half that of Shift8 and Shift8-Dilated. As illustrated in \cref{fig:appendix_shift_step_lam}, we can observe that there are still large activating regions but smaller activating values.

\begin{figure}[htb]
  \centering
 \includegraphics[width=0.9\textwidth]{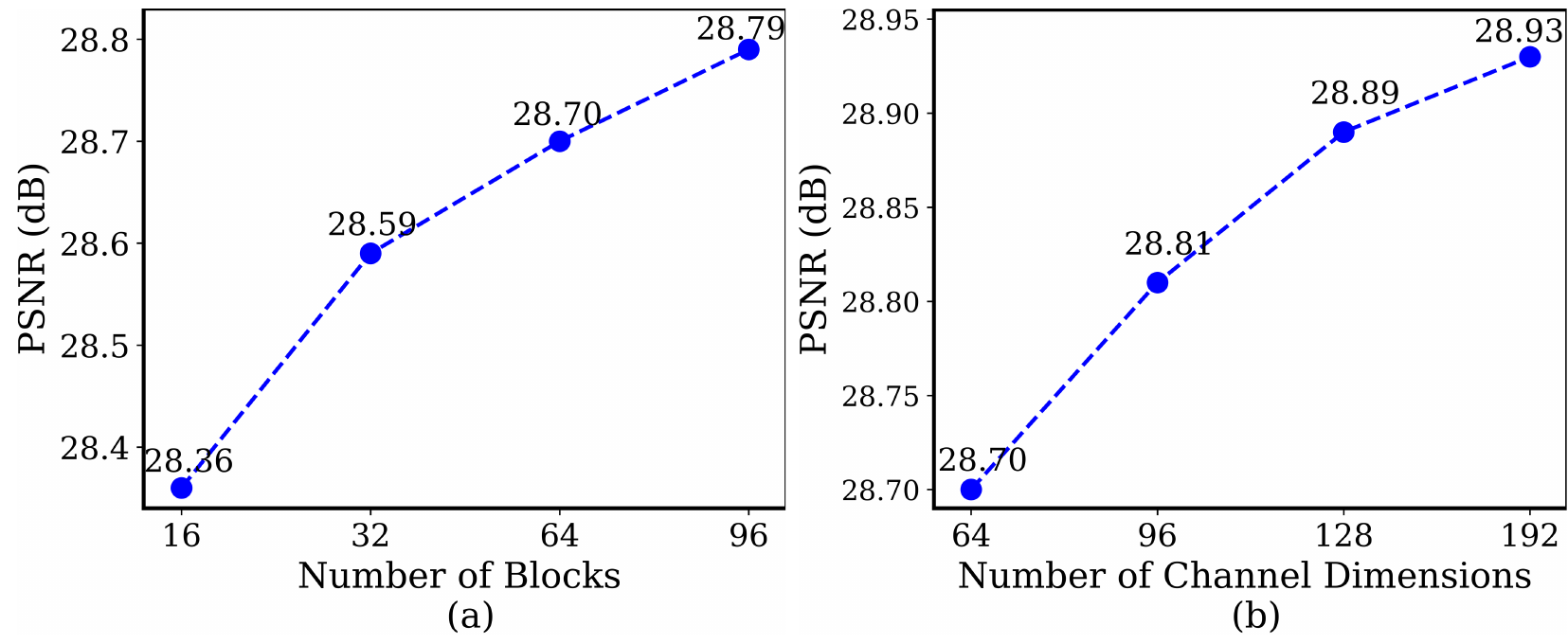}
 \caption{Results of SCNet on Set14 ($\times4$) with different model capacities. (a) Increasing the number of SC-ResBlock with a fixing channel dimension 64. (b) Increasing the number of channel dimension with 64 SC-ResBlocks.\label{apx:ablation_block_dim}}
\end{figure}

\begin{figure}[tb]
  \centering
 \includegraphics[width=0.45\textwidth]{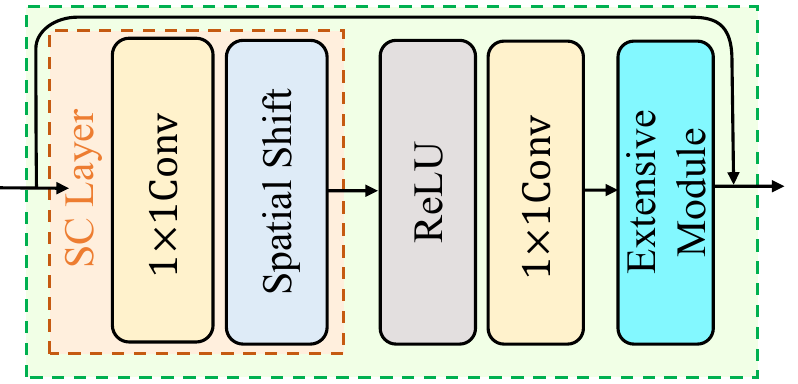}
 \caption{Illustration of the extended SC-ResBlock, and attention modules are obtained by replacing the extensive module. \label{apx:attention}}
\end{figure}

\textbf{The Impact of Model Capacity. \label{sec:ablation:capacity}}
Benefiting from the few parameters in the SC layer, there are opportunities to explore more depths and widths of SCNet. Here we exploit our SCNets stacked with different SC-ResBlocks to analyze the impact of the model capacity. As summarized in \cref{tab:appendix_capacity}, we build our SCNets by SC-ResBlocks with different blocks (simplified as B) and channel dimensions (D).
When comparing SCNets with the same channel dimensions, such as 64 channels, we observe that deeper architectures yield better results. This is further supported by Figure \ref{fig:appendix_arch_lam}, which demonstrates that deeper structures bring larger receptive fields.
When we compare the B64D64 and B16D128, we can find that B64D64 obtains better performance with even fewer parameters. We think it is due to the field of local feature aggregation that B64D64 brings larger receptive fields and much more feature aggregation, while shallow architecture in B16D128 lacks.
In addition, the largest SCNet with B32D128 obtains the best performance. As shown in \cref{fig:appendix_arch_lam}, one can find that more activated pixels are obtained in B32D128 than that in B32D64, which shows that the group dimension is of great significance to the feature aggregation as well. 
The trade-off between the depth and width (group dimension) can be further explored in the future. Moreover, detailed ablations about the deeper architecture and larger channel dimension are shown in \cref{apx:ablation_block_dim}, demonstrating that SCNet is scalable to larger model capacities.

\begin{table}[tb]
\begin{center}

\caption{Results of tiny SCNet with different attention modules on scale 4.-- presents our default SCNet-T without attention module. }\label{tab:ablation_attention}

\renewcommand\arraystretch{1.1}
\resizebox{0.75\textwidth}{!}{

  \begin{tabular}{cccccc}
    \toprule
 \multirow{2}{*}{Attn.}    &\multirow{2}{*}{Params}  & Set14 &B100 & Urban100 & Manga109 \\
  &      &  PSNR/SSIM& PSNR/SSIM& PSNR/SSIM& PSNR/SSIM \\
    \midrule
$-^*$ & 159K & 28.36/0.7764 & 27.39/0.7309 & 25.59/0.7696 & 29.72/0.9000 \\
CA & 188K & 28.40/0.7763 & 27.43/0.7308 & 25.67/0.7716 & 29.84/0.9004 \\
SPA & 179K & 28.45/0.7779 & 27.46/0.7318 & 25.71/0.7727 & 29.95/0.9020 \\
PA & 245K & 28.50/0.7791 & 27.49/0.7329 & 25.81/0.7757 & 30.10/0.9038 \\
    \bottomrule
  \end{tabular}
}

\end{center}
\end{table}

\begin{table*}[tb]
  \caption{Results about the impact of up-scaling modules.}
  \label{tab:ablation_up-scaled}
  \centering
\renewcommand\arraystretch{1.2}
  
  \resizebox{\textwidth}{!}{
  \begin{tabular}{cllccccc}
    \toprule
  \multirow{2}{*}{Scale}&\multirow{2}{*}{Up-Scaling}    &\multirow{2}{*}{Params} & Set5 & Set14 &B100 & Urban100 & Manga109 \\
    %\cmidrule(r){3-7}
       &  &      & PSNR/SSIM & PSNR/SSIM& PSNR/SSIM& PSNR/SSIM& PSNR/SSIM \\
    \midrule
\multirow{4}{*}{$\times2$}
& PixelShuffle & 159K  & 37.85/0.9600 & 33.39/0.9161 & 32.06/0.8981 & 31.50/0.9187 & 38.29/0.9764 \\
& Nearest & 146K   & 37.76/0.9597 & 33.37/0.9151 & 31.99/0.8974 & 31.30/0.9197 & 38.14/0.9760 \\
& Bilinear & 146K   & 37.78/0.9597   &33.31/0.9152   & 32.00/0.8974  & 31.24/0.9193 & 38.12/0.9759 \\
& TConv & 151K   & 37.80/0.9598   & 33.40/0.9153    & 32.02/0.8977  & 31.40/0.9207 & 38.18/0.9761 \\

\bottomrule
  \end{tabular}}
\end{table*}

\begin{table*}[tbp]
\begin{center}
\caption{Quantitative comparison on scale 8.}\label{tab:results_x8}
\renewcommand\arraystretch{1.3}
\resizebox{0.975\textwidth}{!}{%
  \begin{tabular}{llcccccc}
    \toprule
  \multirow{2}{*}{Method} & \multirow{2}{*}{Params}  &\multirow{2}{*}{Flops} & Set5 & Set14 &B100 & Urban100 & Manga109 \\
     &  &    & PSNR/SSIM & PSNR/SSIM& PSNR/SSIM& PSNR/SSIM& PSNR/SSIM \\
\hline
VDSR\cite{VDSR} & 665K & $612.6 \mathrm{G}$ & $25.73 / 0.6743$ & $23.20 / 0.5110$ & $24.34 / 0.5169$ & $21.48 / 0.5289$ & $22.73 / 0.6688$ \\
DRCN\cite{DRCN} & 1,774K & $17,974 \mathrm{G}$ & $25.93 / 0.6743$ & $24.25 / 0.5510$ & $24.49 / 0.5168$ & $21.71 / 0.5289$ & $23.20 / 0.6686$ \\
AWSRN\cite{AWSRN} & 2,348K & ${3 3 . 7 G}$ & ${2 6 . 9 7} / {0 . 7 7 4 7}$ & ${2 4 . 9 9 / 0 . 6 4 1 4}$ & ${2 4 . 8 0 / 0 . 5 9 6 7}$ & ${2 2 . 4 5 / 0 . 6 1 7 4}$ & ${2 4 . 6 0 / 0 . 7 7 8 2}$\\
\textbf{SCNet-B (Ours)} & \textbf{599K} & \textbf{17.7G} & \textbf{27.03}/$\mathbf{0.7770}$ & $\mathbf{25.05}$/$\mathbf{0.6414}$ & $\mathbf{24.83}$/$\mathbf{0.5962}$ & $\mathbf{22.57}$/$\mathbf{0.6204}$ & $\mathbf{24.71}$/$\mathbf{0.7840}$ \\
\bottomrule
\end{tabular}}
\end{center}
\end{table*}

\subsection{Scalability of SCNet}
As we discussed before, the primariy goal of this paper is to propose a new benchmark SR network named SCNet by stacking numerous Shift-Conv layer. By applying spatial-shift operation, SCNet can achieve comparable performance compared to existing advanced methods. To comprehensively explore the potential of the SCNet model, we delve into a thorough evaluation of its scalability through a series of rigorous experiments. This exhaustive analysis allows us to better understand the impacts of our proposed model and demonstrates its remarkable scalability.

\textbf{Extensive Attention Modules. \label{sec:ablation:extension}}
The attention mechanism has been shown to play a crucial role in CNN-based methods, particularly for lightweight models. To this end, we extend the proposed SCNet with channel attention (simplified as CA), spatial attention (SPA), and pixel attention(PA), as illustrated in \cref{apx:attention}, and present the results in \cref{tab:ablation_attention}. We can find that spatial attention, which contains fewer parameters than channel attention, achieves better performance. In general, we can conclude that the proposed SCNet is scalable to attention modules, which can bring further improvement. These results verify that the proposed SCNet based on the vanilla residual block can effectively accommodate attention mechanisms, which highlights the potential for future exploration of well-designed architectures for SCNet.

\begin{figure}[tbp]
 \centering
 \includegraphics[width=0.55\textwidth]{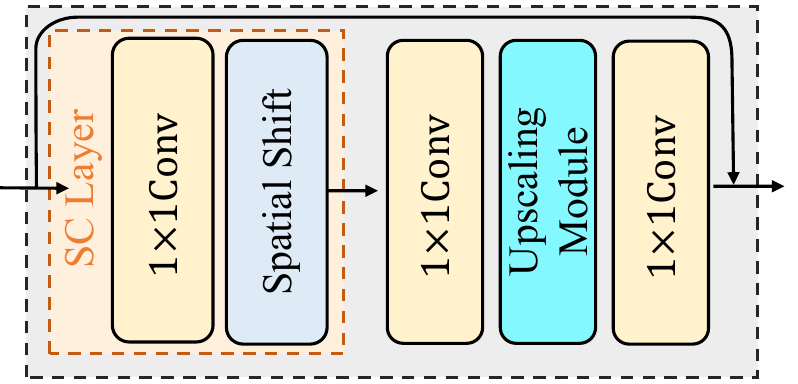}
 \caption{Illustration of the reconstruction block with different up-scaling modules.}
\label{fig:up-scaled}
\end{figure}

\textbf{The Impact of Up-Scaling Modules.}
Unlike traditional CNN-based methods, SCNet exclusively utilizes $1\times1$ convolutions and simplifies the reconstruction module. To assess the adaptability of Shift-Conv to different upscaling strategies, we conducted an investigation using various upscaling approaches. For a fair comparison, we take SCNet-T as the default model and modify the reconstruction module with different up-scaling strategies as illustrated in \cref{fig:up-scaled}. We evaluate four widely utilized up-scaling strategies: transport convolution, convolution with pixelshuffle, bilinear interpolation with convolution, and the nearest interpolation with convolution, which are abbreviated as TConv, PixelShuffle, Bilinear, and Nearest, respectively. Results for $\times 2$ super-resolution are summarized in \cref{tab:ablation_up-scaled}. As shown in \cref{tab:ablation_up-scaled}, the pixelshuffle module with slightly more parameters achieves the best performance on all test datasets. Specifically, SCNet with pixelshuffle obtains 0.10 dB and 0.11 dB improvement on Urban100 and Manga109, respectively, compared to the second-best approach.

\textbf{Extensive SR task.}
To comprehensively evaluate the effectiveness of SCNet, we conducte experiments on a scale factor of 8. The results are presented in \cref{tab:results_x8}, which confirms the efficiency and effectiveness of the proposed SCNet.

In extensive experiments, we enhanced the proposed SCNet with attention mechanisms including channel, spatial, and pixel attention. Applying this extensive module provides an improvement in performance. Additionally, we assessed the impact of various up-scaling modules within SCNet, revealing that our proposed fully $1\times1$ convolution is general to different upscaling approaches. Finally, to further scrutinize its efficacy, we conducted experiments with a larger scaling factor of 8, which provides robust performance, reinforcing its efficiency and effectiveness in high-demand super-resolution tasks.

\subsection{Discussion and Limitation}
In this section, we present both quantitative and qualitative comparisons to showcase the effectiveness of the proposed SCNet. Furthermore, we provide thorough ablations to analyze the impact of various components in SCNet, including the receptive fields, the trade-off between space extension and group dimensions, and extensive modules. These results provide deep insight and indicate that the proposed SCNet has great potential for further study. 

While the proposed SCNet effectively and efficiently addresses lightweight SISR, there are still challenges to be addressed in the future. This paper only explores the vanilla residual connection-based architecture. As presented in \cref{tab:main_results} and \cref{apx:attention}, we believe that well-designed architectures could further enhance the model capabilities, such as the large kernel design in recent CNNs and long-rang modeling in Transformer, but this is beyond the scope of this paper. Additionally, as shown in \cref{fig:appendix_shift_step_lam} and \cref{tab:appendix_shift_step}, SCNet is scalable in obtaining larger receptive fields. However, more complex mechanisms, such as adaptive shift, are meaningful to study in the future.

\section{Conclusion \label{sec:conclusion}}
In this paper, we pivots away from the conventional approach of devising increasingly complex network architectures, and instead opting for a minimalist and fully $1\times1$ convolutional network named SCNet, leading to a marked reduction in both parameters and computational costs. Nonetheless, $1\times1$ convolution brings its own challenges, primarily the absence of local feature aggregation, a critical aspect of effective modeling. To overcome this, we expand the $1\times1$ convolution into the Shift-Conv layer. By incorporating a spatial-shift operation, it facilitates local feature aggregation along the channel dimension without adding computational overhead. Our thorough experiments have demonstrated that SCNet can match or even outperform existing advanced methods. Moreover, in-depth analyses highlights the versatility and scalability of SCNet as a robust baseline architecture. We hope that our work with the SCNet will ignite further exploration in the research community, encouraging the development of advanced local and long-range feature aggregation patterns.

\section*{Acknowledgements}
The research was supported by the National Natural Science Foundation of China (U23B2009, 92270116), and was partially supported by the Fundamental Research Funds for the Central Universities.

\bibliography{mir-bib}

\begin{figure}[h]%
\centering
\includegraphics[width=0.3\textwidth]{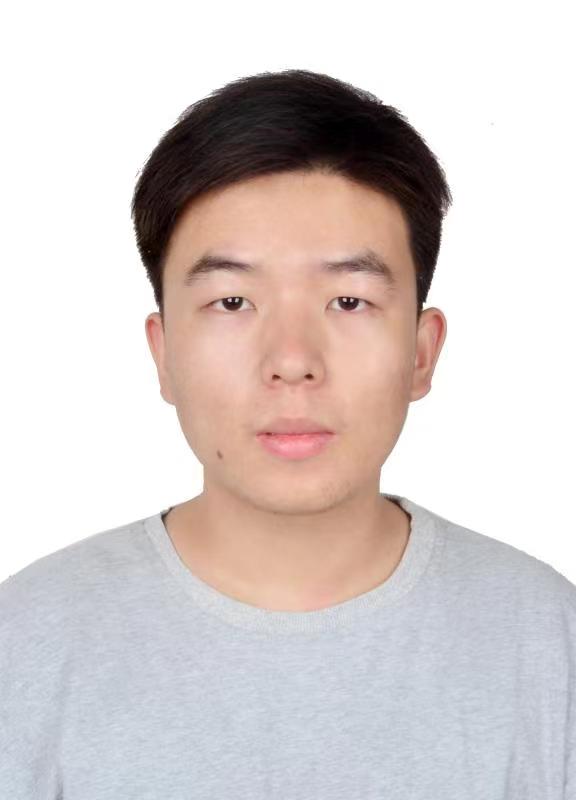}
\end{figure}

\noindent{\bf Gang Wu} received the B.E. degree in the School of Computer Science and Technology from Soochow University, Jiangsu, China, in 2020. He is currently pursuing the Ph.D. degree in Faculty of Computing at Harbin Institute of Technology. His research interests include image restoration, representation learning, and self-supervised learning. E-mail: gwu@hit.edu.cn

ORCID iD: 0009-0007-5003-3117

\begin{figure}[h]%
\centering
\includegraphics[width=0.3\textwidth]{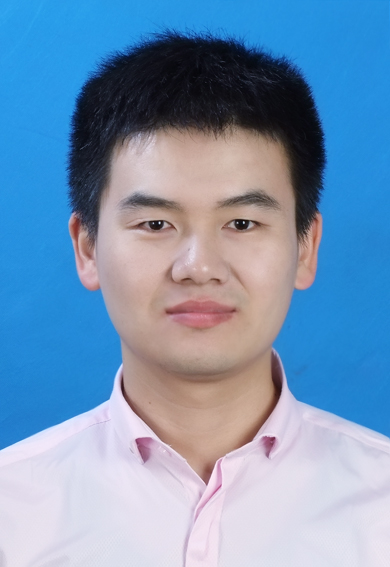}
\end{figure}

\noindent{\bf Junjun Jiang} received the B.S. degree in Mathematics from the Huaqiao University, Quanzhou, China, in 2009, and the Ph.D. degree in Computer Science from the Wuhan University, Wuhan, China, in 2014. From 2015 to 2018, he was an Associate Professor with the School of Computer Science, China University of Geosciences, Wuhan. From 2016 to 2018, he was a Project Researcher with the National Institute
of Informatics (NII), Tokyo, Japan. He is currently a Professor with the School of Computer Science and Technology, Harbin Institute of Technology, Harbin, China. He won the Best Student Paper Runner-up Award at MMM 2017, the Finalist of the World's FIRST 10K Best Paper Award at ICME 2017, and the Best Paper Award at IFTC 2018. He received the 2016 China Computer Federation (CCF) Outstanding Doctoral Dissertation Award and 2015 ACM Wuhan Doctoral Dissertation Award. E-mail: jiangjunjun@hit.edu.cn (Corresponding Author)

ORCID iD: 0000-0002-5694-505X

\begin{figure}[h]%
\centering
\includegraphics[width=0.3\textwidth]{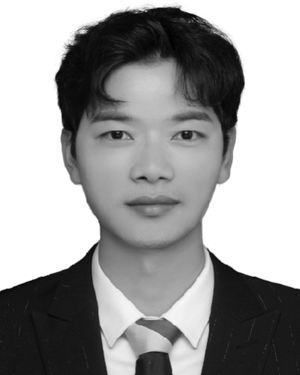}
\end{figure}

\noindent{\bf Kui Jiang} received the M.E. and Ph.D. degrees from the School of Computer Science, Wuhan University, Wuhan, China, in 2019 and 2022, respectively. Before July 2023, he was a Research Scientist with the Cloud BU, Huawei. He is currently an Associate Professor with the School of Computer Science and Technology, Harbin Institute of Technology. He received the 2022 ACM Wuhan Doctoral Dissertation Award. His research interests include image/video processing and computer vision. E-mail: jiangkui@hit.edu.cn

\begin{figure}[h]%
\centering
\includegraphics[width=0.3\textwidth]{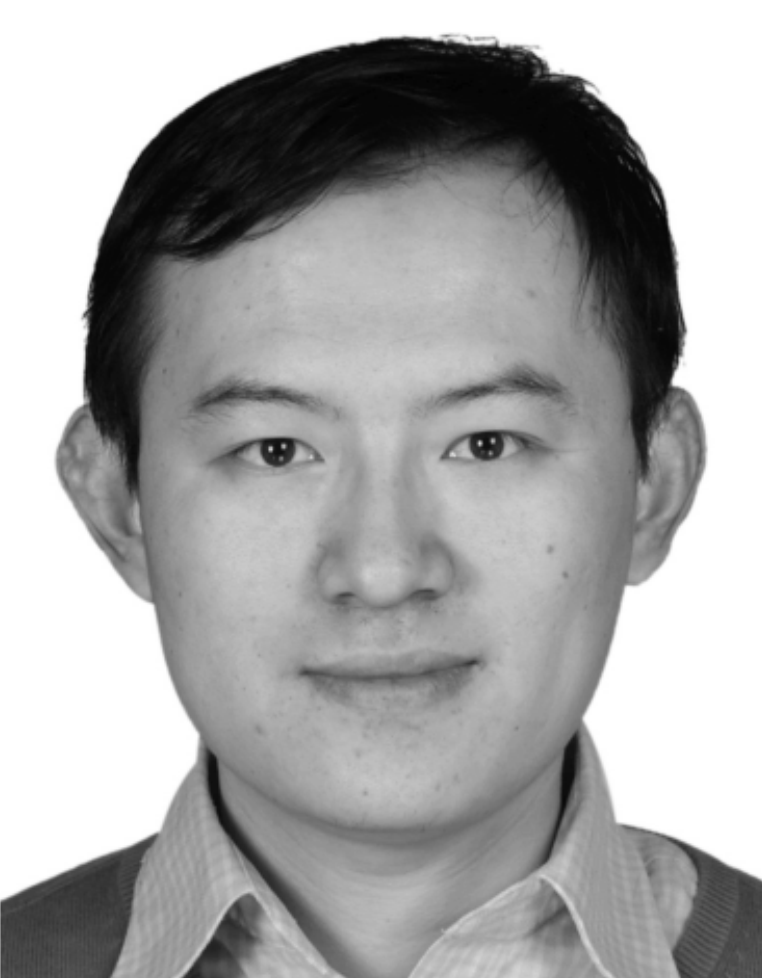}
\end{figure}
\noindent{\bf Xianming Liu} received the B.S., M.S., and Ph.D. degrees in computer science from the Harbin
Institute of Technology (HIT), Harbin, China,
in 2006, 2008, and 2012, respectively. In 2011, he
spent half a year at the Department of Electrical
and Computer Engineering, McMaster University,
Canada, as a Visiting Student, where he was a
Post-Doctoral Fellow from 2012 to 2013. He was
a Project Researcher with the National Institute of
Informatics (NII), Tokyo, Japan, from 2014 to 2017.
He is currently a Professor with the School of Computer
Science and Technology, HIT. He was a receipt of the IEEE ICME 2016 Best Student Paper Award. E-mail: csxm@hit.edu.cn

\end{document}